# SPARNet: Continual Test-Time Adaptation via Sample Partitioning Strategy and Anti-Forgetting Regularization

Xinru Meng[1], Sun Han[1*], Jiamei Liu[1], Ningzhong Liu[1], and Huiyu Zhou[2]

[1]College of Computer Science and Technology
Nanjing University of Aeronautics and Astronautics, Nanjing 210016, China
[2]School of Computing and Mathematical Sciences,
University of Leicester, Leicester LE1 7RH, UK
*Email: sunhan@nuaa.edu.cn

*Abstract*

Test-time Adaptation (TTA) aims to improve model performance when the model encounters domain changes after deployment. The standard TTA mainly considers the case where the target domain is static, while the continual TTA needs to undergo a sequence of domain changes. This encounters a significant challenge as the model needs to adapt for the long-term and is unaware of when the domain changes occur. The quality of pseudo-labels is hard to guarantee. Noisy pseudo-labels produced by simple self-training methods can cause error accumulation and catastrophic forgetting. In this work, we propose a new framework named SPARNet which consists of two parts: sample partitioning strategy and anti-forgetting regularization. The sample partition strategy divides samples into two groups, namely reliable samples and unreliable samples. According to the characteristics of each group of samples, we choose different strategies to deal with different groups of samples. This ensures that reliable samples contribute more to the model. At the same time, the negative impacts of unreliable samples are eliminated by the mean teacher's consistency learning. Finally, we introduce a regularization term to alleviate the catastrophic forgetting problem, which can limit important parameters from excessive changes. This term enables long-term adaptation of parameters in the network. The effectiveness of our method is demonstrated in continual TTA scenario by conducting a large number of experiments on CIFAR10-C, CIFAR100-C, and ImageNet-C.

## I. Introduction

The significant advancements in deep learning have demonstrated its importance. However, the majority of current deep learning approaches are grounded in the assumption that training and test data are independent and identically distributed. Unfortunately, this assumption often fails in real-world scenarios, resulting in a potential degradation of model performance when faced with new distribution test data. To solve the issues mentioned above, the representative method is domain adaptation which aims to transfer the knowledge learned from the source domain data to the target domain data. Furthermore, model adaptation is necessary in an online manner as distribution shifts can occur after the model deployment. Consequently, test-time adaptation (TTA) methods have widely received research interests. TTA uses unlabeled test data streams to dynamically adjust a pre-trained model based on the source data, ensuring it adapts to the current data distribution during test-time. To provide a more realistic representation of challenges encountered in real-world scenarios, a continual test-time adaptation approach [1] was recently proposed which requires the model to adapt to the multiple domain shifts. This undoubtedly increases the difficulty of the task.

Numerous studies have shown the effectiveness of various methods in dealing with a single domain shift. One such approach is adjusting batch normalization statistics [2] during test-time, which can already significantly improve the performance. Another approach updates model weights through self-training methods, such as entropy minimization [3]. [4,5] alleviate error accumulation by filtering out unreliable samples. However, we argue that unreliable samples with low-confidence still contain valuable underlying distribution information. Nevertheless, when faced with continual domain changes and the need for long-term model adaptation, the aforementioned methods inevitably encounter great limitations. Constantly updating the model can result in the problem of catastrophic forgetting [6-8] and the loss of pre-trained model knowledge.

To address the aforementioned challenges, we propose a sample partitioning strategy for handling newly arrived target domain data. This strategy divides test samples into reliable and unreliable groups based on the model's prediction outputs. For reliable samples, an extended entropy minimization loss is introduced to generate large gradients and more positive updates of the model during test-time. On the other hand, due to unreliable samples are associated with highly uncertain predictions, we propose mean teacher consistency loss to promote the feature representation learning by self-supervised learning. Different augmented views of one sample should have similar predictions between the student model and the teacher model. In addition, the source data is unavailable due to privacy concerns. The model continuously accumulates new knowledge under long-term adaptation, leading to catastrophic forgetting. To address this, we use an anti-forgetting regularization term which ensures the model preserve important information of the initial task when learning new domain data. By implementing these methods, we aim to improve the model's adaptability to continual domain shifts, mitigate the impact of unreliable samples, and prevent catastrophic forgetting when accumulating new knowledge over time.

Our main contributions can be summarized as follows:
1) We propose a new sample partitioning strategy that divides the target test samples into two groups and uses different strategies to study the two groups of samples. This strategy alleviates the problem of error accumulation and improves the model's performance in handling continual domain changes.
2) To maintain stability and the long-term learning capability of the model, we introduce a label-independent weight regularization term to prevent drastic changes in important model weights. This term uses the gradient of the output function rather than the gradient of the loss function.
3) The proposed approach demonstrates the performance improvements in continual test-time adaptation tasks on various classification benchmarks.

## II. RELATED WORK

### A. Unsupervised Domain Adaptation(UDA)

The earliest unsupervised domain adaptation method(UDA) uses labeled source domain data and unlabeled target domain data to deal with the domain shift problem. In these initial UDA frameworks, the focus is primarily on reducing the impact of domain transfer through instance re-weighting techniques [9-11]. The objective is to improve the model's performance in the target domain by re-weighting source data instances, thereby making the source distribution closer to the target distribution in a non-parametric manner and reducing the discrepancy between the two domains. Other research focuses on using deep learning techniques to minimize the discrepancy between source and target domain features. This objective was achieved through various methods such as adversarial learning [12,13], discrepancy based loss functions [14-16], or contrastive learning [17,18] to achieve. However, it's worth noting that the aforementioned methods require access to the source data, which is not only inefficient but may also violate privacy or access restrictions.

### B. Continual Test-time Adaptation

During test-time adaptation, the model needs to be adjusted in real time to fit the target data that has distribution shift from the source domain data. Updating the batch normalization (BN) statistics using the target data during test-time shows encouraging results [2]. LAME [19] uses Laplacian adjusted maximum-likelihood estimation to adapt the model's outputs. Both of them do not involve model parameters updates. Most existing methods often update the model weights by self-training in the test time adaptation task. TENT [3] minimizes the entropy of the prediction by updating the BN parameters and performs standard back propagation for model weights updates. Considering the unreliability of high-entropy samples, the loss is only computed based on reliable samples [5]. Additionally, it uses EWC [20] regularization to maintain the stability of the model. Similar to EATA [5], SAR [21] removes noisy samples with large gradients to ensure that the model weights are optimized to a flat minimum.

Initially, the test-time adaptation approach adapts mainly for a single target domain. [1] is the first study to consider continual domain changes. It introduces a mean teacher framework to refine pseudo-labels along with stochastic restoration to prevent catastrophic forgetting. One method utilizes contrastive learning and uses symmetric cross-entropy as the consistency loss of mean teacher framework [22]. Moreover, the different augmented views of one sample are generated to learn transformation-invariant mapping [23]. All above methods further reduce error accumulation by introducing self-supervised learning. In AR-TTA [24], a small memory combined with mixup data augmentation buffer is used to increase model stability. In contrast to these methods, RoTTA [25] additionally takes into account the fact that the real-world data is often temporally correlated and proposes a robust approach for this scenario.

### C. Continual Learning

Continual Learning [26] focuses on models learning from new data and adapting to changing environments over time. During the continuous learning and adaptation process of the model, previously learned knowledge is difficult to preserve because new data will interfere with the model. In order to adapt to new tasks, the model adjusts the parameters which have been learned from previous data, resulting in the forgetting of previous knowledge. Some continual learning methods utilize replay-based methods [27], which store previously learned knowledge and use it to guide the new data learning process. This can help the model avoid forgetting previously learned information and improve its ability to adapt to new data and environments. However, the methods can be computationally expensive and difficult to implement in practice. There are other methods based on regularization [28,29], which limit the complexity of the model by imposing penalties on the parameters of the model, thereby making the model more general and better able to cope with new data and changing environments. For example, learning without forgetting (LwF) [30], elastic weight consolidation(EWC) [20] and MAS [31].

## III. METHOD

### A. Problem Definition and SPARNet framework

The continual test-time adaptation task continuously adapt to the changing target domain in an online manner without using source domain data during test-time, improving the performance of the model pre-trained on the source domain data. That is to say, given a model $f_{\theta_0}$ pre-trained on the source domain $D^s = \{x_n^s, y_n^s\}_{n=1}^{N^s}$, the unlabeled target domain data $X^T$ will be arrived sequentially. The distribution of the target domain changes with time. At time step $t$, the model $f_{\theta_t}$ can only access the data $x_t^T$ of the current time step and makes predictions immediately after the data arrives. The parameters $\theta_t \to \theta_{t+1}$ are then adjusted accordingly.

The overall architecture of our proposed SPARNet framework is shown in Fig. 1. SPARNet uses a student network to divide the samples into two groups based on entropy which is calculated by student model predictions. Reliable samples learn target features through self-training loss to produce large contribution to the model. By constructing different views of unreliable samples, consistency loss is performed between the student model and the teacher model to alleviate error accumulation. Finally, we introduce an anti-forgetting regularization term to avoid catastrophic forgetting and maintain the stability of the model.

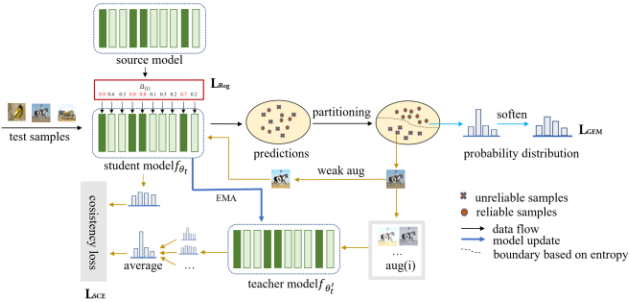

Fig. 1. The framework of our proposed SPARNet method. The samples first are divided into two groups: reliable samples and unreliable samples. We soften the probability distribution of reliable samples, while unreliable samples are utilized for consistency learning across views. Parameters importance obtained from the source model is used to limit the update of student model weights.

### B. Target samples partitioning based on entropy

In order to perform effective test-time adaptation and reduce the error accumulation problem, we propose a sample partitioning strategy to promote model updating. Specifically, despite the existence of domain shift, the test samples will have different effects in adaptation as their different degree of shift or recognition difficulty. In other words, high-confidence samples should give different feedback to the model than low-confidence samples. Therefore, for the incoming target sample $x_t^T$, we use the model $f_{\theta_t}$ to obtain the predicted probability outputs of samples. By calculating the entropy of samples, we divide the samples of each mini-batch data into two groups: the unreliable samples and the reliable samples:

$$x_{tr}^T = \{x \mid x \in x_t^T, H(y_t^T \mid x) < E_0\},$$
$$x_{tu}^T = \{x \mid x \in x_t^T, H(y_t^T \mid x) \geq E_0\}. \quad (1)$$

where $y_t^T = \delta(z_t^T = f_{\theta_t}(x_t^T))$ is the probability output ($\delta$ represents the softmax operation) and $H(y_t^T) = -\sum_{c=1}^C y_{tc}^T \log y_{tc}^T$ is the entropy of the sample, where C is the number of classes. $E_0$ is a pre-defined threshold.

We use the threshold $E_0$ to divide the samples into two groups. Reliable samples with low entropy value produce large contribution for weight updating to the model. While for unreliable high entropy samples, we use self-supervised learning to obtain additional feature information towards the unknown target domain distribution and help the model learn meaningful representations from these challenging samples.

### C. Generalized Entropy Minimization

In unsupervised learning, where sample labels are unavailable, entropy minimization (EM) loss has been widely used due to its simplicity effectiveness [3].

$$\mathcal{L}_{\text{EM}}(x_{tr}^T) = -\sum_{c=1}^C y_{trc}^T \log y_{trc}^T. \quad (2)$$

where $C$ is the number of classes and $y_{tr}^T = \delta(z_{tr}^T = f_{\theta_t}(x_{tr}^T))$ represents the predicted output of the student model.

The model has identified samples with high level of reliability. [32] has found that when adjusting the network with entropy minimization as the loss function, these samples that are highly predicted by the model tend to have less impact on the weight adjustment. Therefore, we use a new generalized entropy minimization loss (GEM) [32] in order to get more positive feedback and more larger loss from the reliable high-confidence samples. The loss function increases its contribution to the model by softening the logits distribution to make the distribution smoother. In this way, we introduce higher loss values for these well-predicted samples. This enables the model to learn and adapt more efficiently:

$$\mathcal{L}_{\text{GEM}}(x_{tr}^T) = -\tau^2 \sum_{c=1}^C y_{trc}^T \log y_{trc}^T, \quad y_{tr}^T = \frac{\exp(z_{tr}^T/\tau)}{\sum_{c=1}^C \exp(z_{trc}^T/\tau)}. \quad (3)$$

We soften the logits distribution by making $\tau \geq 1$. We follow previous work [33] and set it to $\tau = \frac{s}{N} \sum_{i=1}^N \sigma_{z_i^T}$, where 4s is a hyperparameter used to adjust the scaling strength and set to 1. That is to say, the logits are scaled using the standard deviation of the logits $\sigma_{z_t^T}$ as the dynamic temperature.

### D. Mean Teacher Consistency

In recent some studies, [5,21] drop all high entropy samples. However, we argue that discarding all high entropy samples may result in the loss of useful information. Simply filtering the data will ignore the underlying distribution information related to the target domain features, which will affect the model's feature learning. It is reasonable that the model gives low-confidence predictions for some samples. After all, there exists domain shifts in the data distribution. When facing samples with low confidence, traditional self-training methods may effect the performance and generalization ability of the model due to the unreliability of the model outputs.

Self-updating using network-generated pseudo-labels has been shown to be an extremely effective method for the test-time adaptation task. However, considering the unreliability of high-entropy samples, we use the more robust augmentation-average pseudo-labels from [1] instead of network-generated pseudo-labels. Specifically speaking, we use the source model obtained by training on the source domain data to initialize two identical models: student model $f_{\theta_t}$ and teacher model $f_{\theta'_t}$. They have the same weights initially, both of which are the source model training weights. The teacher network computes the average of the predictions obtained from the different augmented versions of the sample, which enables teacher network to provide a more refined and robust pseudo-labels.

The student model $f_{\theta_t}$ is updated by the loss of cross-entropy( CE ) between student predictions and the robust pseudo-labels obtained by the teacher.

$$\widehat{y}_{tu}^T = \frac{1}{N} \sum_{i=0}^{N-1} f_{\theta'_t}(\text{aug}_i(x_{tu}^T)). \quad (4)$$

$$\mathcal{L}_{\text{CE}}(x_{tu}^T) = -\sum_{c=1}^{C} \widehat{y}_{tuc}^T \log y_{tuc}^T. \quad (5)$$

where $\widehat{y}_{tu}^T$ and $y_{tu}^T$ are the augmentation-averaged predictions of the teacher model and the predictions of the student model for the unreliable sample, respectively.

Then, the weights $\theta'$ of the teacher model are updated by exponential moving average based on the student's weights $\theta$:

$$\theta'_{t+1} = \alpha \theta'_t + (1 - \alpha) \theta_{t+1}. \quad (6)$$

where $\alpha$ is a smoothing factor.

Following the manifold smoothness assumption in the field of semi-supervised learning( SSL ) [34-37], the output of the model should not change when giving the data points an realistic perturbation [38]. To take advantage of potentially beneficial supervisory signals from unlabeled data and maintain consistency between teacher and student under smaller perturbations, weak augmentation is applied to low-confidence samples. Weak augmentation operations are simple random cropping and flipping of images. The weak augmented view is predicted by the student network. Ultimately, we will promote consistency between the robust pseudo-labels and the classification output under the weak augmented view.

When training the model with standard cross-entropy( CE ) loss, there usually exists issues of overfitting for simple classes and underfitting for difficult classes. The method proposed by [22] argues that the symmetric cross-entropy( SCE ) [39] loss has better gradient properties compared with the standard cross-entropy loss when using the mean teacher framework. Symmetric cross-entropy loss can be mathematically expressed as follows:

$$\mathcal{L}_{\text{SCE}}(x_{tu}^T, x_{tu}^{'T}) = \frac{1}{2} \mathcal{L}_{\text{CE}}(\widehat{y}_{tu}^T, y_{tu}^{'T}) + \mathcal{L}_{\text{CE}}(y_{tu}^{'T}, \widehat{y}_{tu}^T) \quad (7)$$

where $x_{tu}^{'T} = \Lambda_w(x_{tu}^T)$ is weak augmentation view of $x_{tu}^T$ and $y_{tu}^{'T}$ is the student model's prediction for that view.

We ensemble the student model predictions and the teacher's augmentation-average predictions. The reason is that the stude nt model learns more features of the target data and the more stable teacher model extracts domain-invariant features as multifaceted as possible. The ensemble outputs of the student-teacher get a dual representation of the feature output of the target data.

*E. Anti-Forgetting Regularization*

In the continual test-time adaptation task, the model encounters continual domain changes and updates the model weights frequently. The noise generated under long-term adaptation introduces errors and leads to model forgetting. To alleviate the above issues, we preserve the model's initial knowledge and avoid model performance degradation by introducing an anti-forgetting regularization term [31]. This regularization term is co-optimized with the other losses of the model:

$$\mathcal{L}_{\text{Reg}}(\theta_t, \theta_0) = \sum_{\theta_i \in \theta_t} \Omega(\theta_i)(\theta_i - \theta_{0i})^2. \quad (8)$$

where $\theta_t$ are parameters used for model update and $\theta_0$ are the corresponding parameters of the original model. $\Omega(\theta_i)$ denotes the importance of $\theta_i$. $\Omega(\theta_i)$ penalizes large changes in important parameters when adapting to a new distribution. Besides, parameters with low importance can still be used to optimize the function for new tasks.

We use the sensitivity of the output function to estimate the parameter importance $\Omega(\theta_i)$ [31]. Regularization-based methods inevitably use the original distribution samples. However, we only need to compute $\Omega(\theta_i)$ once before the model deployment. During test-time, $\Omega$ is fixed and the original distribution samples are no longer used:

$$\Omega(\theta_i) = \frac{1}{Q} \sum_{q \in Q} \left( \left| \frac{\partial [\ell_2^2 f_{\theta_0}(x_q)]}{\partial \theta_{0i}} \right| \right). \quad (9)$$

where $\ell_2$ is the $L_2$ norm and $Q$ denotes the source domain data set. The parameter corresponding to the large value in $\Omega(\theta_i)$ is minimized by the magnitude of its change during gradient descent. As the parameter is important for past tasks, the parameter value needs to be preserved.

Therefore, our overall optimization objective is:

$$\mathcal{L} = \mathcal{L}_{\text{SCE}} + \lambda \mathcal{L}_{\text{GEM}} + \beta \mathcal{L}_{\text{Reg}}. \tag{10}$$

where $\lambda$ and $\beta$ are hyperparameters.

## IV. EXPERIMENTS

### A. Experimental setup

1) Dataset

We illustrate the effectiveness of our method on three commonly benchmark datasets, including ImageNet-C, CIFAR10-C, and CIFAR100-C. These datasets include 15 different types of image corruption with 5 levels of severity. We use a continual benchmark [1]. The model is adapted to a sequence of test domains in an online manner. We evaluate the model at the largest corruption severity level of 5.

2) Implementation details

To ensure a fair comparison with related methods, we follow other state-of-the-art continual test-time adaptation techniques. In all experiments, we use WideResNet-28 [40], ResNeXt-29 [41] and the uniformly pre-trained ResNet50 model in the CIFAR10 to CIFAR10-C, CIFAR100 to CIFAR100-C, and ImageNet to ImageNet-C tasks, respectively, all of which are taken from the RobustBench benchmark [42]. For CIFAR10-C and CIFAR100-C datasets, we using the Adam optimizer with a learning rate of 1e-3 and the batch size is set to 200. For ImageNet-C dataset, we using the SGD optimizer with a learning rate of 0.0002 and the batch size is set to 32.

Regarding the hyperparameter settings of our method, $E_0$ is set to $0.4 \times \ln C$ for all datasets by following EATA [5], where $C$ is number of dataset classes. Following [1], we use 32 stochastic augmentation in our experiments. The value of $\lambda$ is set to 1.8, 1.8, and 0.3 for CIFAR10-C, CIFAR100-C and ImageNet-C, respectively. The trade-off parameter $\beta$ is set to 1, 1, and 0.005 for CIFAR10-C and CIFAR100-C/ImageNet-C, respectively. We use 2000 samples to calculate $\Omega$.

3) Baselines

We compared our method with several state-of-the-art methods, including source-only: Source, BN Adapt [2] (NIPS 2020), TENT [3] (ICLR 2021), LAME [19] (CVPR 2022), CoTTA [1] (CVPR 2022), RoTTA [25] (CVPR 2023), AR-TTA [24] (ICCV 2023).

### B. Results

The experimental outcomes of the three datasets are presented in Table I-III. Using the most original pre-trained model directly without adaptation, the average error rate performs very poorly on all datasets. The BN Adapt method, which does not update the model weights but adapts based on the batch normalization statistics of the current input data, improves performance significantly. The TENT-cont. method hopes to improve performance by updating the model weights, but suffers from severe error accumulation. And the result in CIFAR100-C is even worse than using the source model, rising to 60.9%. Compared to the three most effective methods, our method performs well on all three tasks with average error rates of 16.0%, 30.6%, and 67.3%. The sample partitioning strategy enables the model to learn the target data features more easily and adequately from reliable samples and acquire more underlying distribution information by self-supervised learning of unreliable samples. At the same time, anti-forgetting regularization term alleviates the catastrophic forgetting problem and enhances the learning ability and stability of the model. These results confirm the effectiveness of our approach.

TABLE I
CLASSIFICATION ERROR RATE (%) FOR THE CIFAR10-TO-CIFAR10-C CONTINUAL TEST-TIME ADAPTATION TASK. RESULTS ARE EVALUATED ON THE LARGEST CORRUPTION SEVERITY LEVEL 5

$t \longrightarrow$

| method | Gau. | Shot | Imp. | Def. | Gla. | Mot. | Zoom | Snow | Fro. | Fog | Bri. | Con. | Ela. | Pix. | Jpeg | Mean |
|---|---|---|---|---|---|---|---|---|---|---|---|---|---|---|---|---|
| Source | 72.3 | 65.7 | 72.9 | 46.9 | 54.3 | 34.8 | 42.0 | 25.1 | 41.3 | 26.0 | 9.3 | 46.7 | 26.6 | 58.5 | 30.3 | 43.5 |
| BN Adapt | 28.1 | 26.1 | 36.3 | 12.8 | 35.3 | 14.2 | 12.1 | 17.3 | 17.4 | 15.3 | 8.4 | 12.6 | 23.8 | 19.7 | 27.3 | 20.4 |
| TENT-cont. | 24.8 | 20.6 | 28.6 | 14.4 | 31.3 | 16.5 | 14.1 | 19.1 | 18.6 | 18.6 | 12.2 | 20.3 | 25.7 | 20.8 | 24.9 | 20.7 |
| LAME | 86.0 | 83.9 | 88.4 | 83.6 | 88.7 | 64.4 | 82.0 | 28.4 | 71.7 | 37.1 | 9.4 | 74.1 | 41.3 | 79.7 | 46.3 | 64.3 |
| CoTTA | 24.3 | 21.3 | 26.6 | **11.6** | 27.6 | **12.2** | 10.3 | 14.8 | 14.1 | **12.4** | **7.5** | **10.6** | **18.3** | **13.4** | **17.3** | 16.2 |
| RoTTA | 30.3 | 25.4 | 34.6 | 18.3 | 34.0 | 14.7 | 11.0 | 16.4 | 14.6 | 14.0 | 8.0 | 12.4 | 20.3 | 16.8 | 19.4 | 19.3 |
| AR-TTA | 30.8 | 25.2 | 33.6 | 15.5 | 32.2 | 16.3 | 14.8 | 18.6 | 17.3 | 16.6 | 12.0 | 15.3 | 26.1 | 21.4 | 23.0 | 21.2 |
| ours | **21.6** | **17.5** | **24.9** | 12.3 | **27.6** | 12.9 | **10.3** | **14.5** | **13.6** | 12.7 | 7.7 | 11.6 | 19.8 | 14.7 | 18.9 | **16.0** |

TABLE II
CLASSIFICATION ERROR RATE (%) FOR THE CIFAR100-TO-CIFAR100-C CONTINUAL TEST-TIME ADAPTATION TASK. RESULTS ARE EVALUATED ON THE LARGEST CORRUPTION SEVERITY LEVEL 5

$t \longrightarrow$

| method | Gau. | Shot | Imp. | Def. | Gla. | Mot. | Zoom | Snow | Fro. | Fog | Bri. | Con. | Ela. | Pix. | Jpeg | Mean |
|---|---|---|---|---|---|---|---|---|---|---|---|---|---|---|---|---|
| Source | 73.0 | 68.0 | 39.4 | 29.3 | 54.1 | 30.8 | 28.8 | 39.5 | 45.8 | 50.3 | 29.5 | 55.1 | 37.2 | 74.7 | 41.2 | 46.4 |
| BN Adapt | 42.1 | 40.7 | 42.7 | 27.6 | 41.9 | 29.7 | 27.9 | 34.9 | 35.0 | 41.5 | 26.5 | 30.3 | 35.7 | 32.9 | 41.2 | 35.4 |
| TENT-cont. | 37.2 | 35.8 | 41.7 | 37.9 | 51.2 | 48.3 | 48.5 | 58.4 | 63.7 | 71.1 | 70.4 | 82.3 | 88.0 | 88.5 | 90.4 | 60.9 |
| LAME | 98.9 | 99.0 | 98.2 | 98.1 | 98.8 | 98.1 | 98.0 | 98.2 | 98.8 | 98.9 | 98.0 | 98.9 | 98.1 | 99.0 | 98.4 | 98.5 |
| CoTTA | 40.1 | 37.7 | 39.7 | 26.9 | 38.0 | 27.9 | 26.4 | 32.8 | 31.8 | 40.3 | 24.7 | **26.9** | 32.5 | 28.3 | **33.5** | 32.5 |
| RoTTA | 49.1 | 44.9 | 45.5 | 30.2 | 42.7 | 29.5 | 26.1 | 32.2 | 30.7 | 37.5 | 24.7 | 29.1 | 32.6 | 30.4 | 36.7 | 34.8 |
| ours | **36.9** | **33.5** | **33.2** | **25.3** | **36.5** | **27.1** | **25.1** | **30.6** | **30.5** | **35.6** | **23.8** | 28.0 | **32.0** | **27.0** | 35.0 | **30.6** |

TABLE III
CLASSIFICATION ERROR RATE (%) FOR THE IMAGENET-TO-IMAGENET-C CONTINUAL TEST-TIME ADAPTATION TASK. RESULTS ARE EVALUATED ON THE LARGEST CORRUPTION SEVERITY LEVEL 5

$t \longrightarrow$

| method | Gau. | Shot | Imp. | Def. | Gla. | Mot. | Zoom | Snow | Fro. | Fog | Bri. | Con. | Ela. | Pix. | Jpeg | Mean |
|---|---|---|---|---|---|---|---|---|---|---|---|---|---|---|---|---|
| Source | 95.3 | 94.5 | 95.3 | 84.8 | 91.0 | 86.8 | 77.1 | 84.3 | 79.7 | 77.2 | 44.4 | 95.5 | 85.2 | 76.9 | 66.6 | 82.3 |
| BN Adapt | 88.0 | 88.0 | 88.1 | 88.9 | 87.6 | 78.7 | 66.4 | 68.5 | 71.0 | 56.3 | 37.3 | 89.8 | 59.8 | 57.7 | 68.0 | 72.9 |
| TENT-cont. | 84.2 | **78.0** | **76.4** | 81.9 | 80.4 | 74.2 | 64.0 | 70.6 | 71.8 | 62.9 | 48.3 | 86.0 | 65.2 | 61.1 | 68.4 | 71.6 |
| LAME | 99.9 | 99.9 | 99.9 | 83.6 | 99.8 | 99.8 | 96.7 | 99.9 | 98.7 | 99.8 | 41.6 | 99.7 | 99.9 | 98.3 | 84.3 | 93.5 |
| CoTTA | 87.0 | 82.9 | 78.2 | 81.8 | **78.4** | 71.0 | 64.3 | 67.4 | 67.1 | 60.6 | 53.8 | **70.3** | 60.3 | 57.1 | 60.1 | 69.3 |
| RoTTA | 88.4 | 83.0 | 82.1 | 91.5 | 83.5 | 72.9 | 59.9 | 67.5 | **64.6** | 53.9 | **35.1** | 74.5 | **54.5** | 48.4 | 52.9 | 67.5 |
| AR-TTA | - | - | - | - | - | - | - | - | - | - | - | - | - | - | - | 68.0 |
| ours | **84.2** | 79.0 | 77.4 | **81.8** | 79.7 | **70.9** | 59.8 | **65.5** | 66.9 | **53.6** | 41.1 | 80.8 | 57.9 | 52.3 | 59.5 | **67.3** |

## C. Ablation studies

1) Component analysis

To investigate the effectiveness of the proposed modules, the quantitative results of the model with different components are shown in Table IV. The first row shows the results from the source-only model, used as a baseline for our comparison. By introducing GEM loss for reliable samples, the model learns more feature representations and equips to discriminate between different classes. Our approach reduces the error rate by 8.9% compared to using only the source model. Next, the mean teacher's self-supervised consistency learning alleviates error accumulation of unreliable samples. The error rate is further reduced. Finally, anti-forgetting regularization enables the model not to suffer performance degradation under the long-term adaptation. Our approach achieves an average error of 16.0% when combining all three modules.

TABLE IV
ABLATION OF THE LOSSES ON CIFAR10-C. $\mathcal{L}_{GEM}(wp)$ INDICATES THAT THE GEM LOSS IS APPLIED TO THE WHOLE BATCH

| Backbone(Source) | $\mathcal{L}_{GEM}$ | $\mathcal{L}_{SCE}$ | $\mathcal{L}_{Reg}$ | Avg. |
|---|---|---|---|---|
| ✓ | | | | 43.5 |
| ✓ | ✓ | | | 30.6 |
| ✓ | ✓ | ✓ | | 27.3 |
| ✓ | ✓ | | ✓ | 16.2 |
| ✓ | ✓ | ✓ | ✓ | 16.0 |

2) Hyperparameters

In this subsection, we evaluate the sensitivity of hyperparameters with the tasks CIFAR10-to-CIFAR10-C Fig.

2 and Fig. 3 show the sensitivity of performance for values of $\lambda$ of [1.2,1.4,1.6,1.8,2,2.2,2.4] and values of $\beta$ of [0.7,0.8,0.9,1,1.1,1.2,1.3]. The results show that our method is insensitive to both parameters.

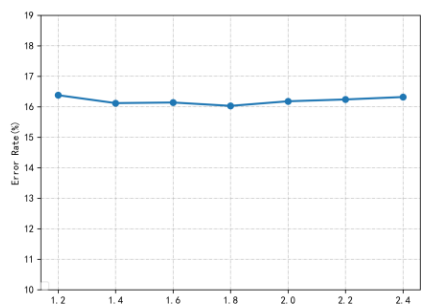

Fig. 2.    Results of parameter λ sensitivity.

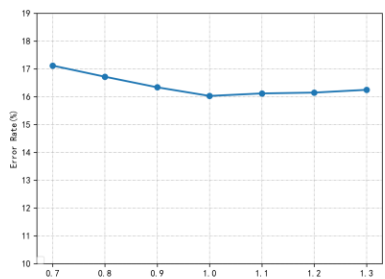

Fig. 3.    Results of parameter β sensitivity.

## V. CONCLUSION

In this paper, we propose a new framework that effectively alleviates the problems of error accumulation and catastrophic forgetting in the continual test time adaptation task. We divide the data into two groups. The high-confidence reliable samples contribute more to the model. The mean teacher consistency loss reduces the negative impact of low-confidence samples. In addition, in order to maintain the stability of model, we use anti-forgetting regularization term to prevent excessive change in important parameters. These methods are integrated to form the SPARNet method, which is validated through experiments on three datasets.